\documentclass[final]{l4dc2024}

\title[Pontryagin Neural Operator for Solving Parametric General-Sum Differential Games]{Pontryagin Neural Operator for Solving General-Sum Differential Games with Parametric State Constraints}
\usepackage{times}
\usepackage{adjustbox}
\usepackage{tablefootnote}

\author{
 \Name{Lei Zhang}\footnotemark[2] \Email{lzhan300@asu.edu}\\
 \Name{Mukesh Ghimire}\footnotemark[2] \Email{mghimire@asu.edu}\\
 \Name{Zhe Xu}\footnotemark[2] \Email{xzhe1@asu.edu}\\
 \Name{Wenlong Zhang}\footnotemark[3] \Email{wenlong.zhang@asu.edu}\\
 \Name{Yi Ren}\footnotemark[2] \Email{yiren@asu.edu}\\
 \addr\footnotemark[2]  Department of Mechanical and Aerospace Engineering, Arizona State University, Tempe, AZ, 85287, USA 
 \addr\footnotemark[3] School of Manufacturing Systems and Networks, Arizona State University, Mesa, AZ, 85212, USA.
}

\usepackage[utf8]{inputenc}
\usepackage{amssymb}
\usepackage{amsmath}
\usepackage{graphicx}  
\usepackage{xcolor}
\DeclareMathOperator*{\argmax}{arg\,max}

\usepackage{wrapfig}   
\usepackage{mathtools}
\usepackage{booktabs}
\usepackage{dsfont}
\usepackage{bbm}
\usepackage{caption}
\usepackage{multirow}
\usepackage[font=footnotesize]{caption} 
\usepackage{pifont} 

\begin{document}

\maketitle
\vspace{-0.15in}
\begin{abstract}
The values of two-player general-sum differential games are viscosity solutions to Hamilton-Jacobi-Isaacs (HJI) equations. Value and policy approximations for such games suffer from the curse of dimensionality (CoD). Alleviating CoD through physics-informed neural networks (PINN) encounters convergence issues when differentiable values with large Lipschitz constants are present due to state constraints. On top of these challenges, it is often necessary to learn generalizable values and policies across a parametric space of games, e.g., for game parameter inference when information is incomplete. To address these challenges, we propose in this paper a Pontryagin-mode neural operator that outperforms the current state-of-the-art hybrid PINN model on safety performance across games with parametric state constraints. Our key contribution is the introduction of a costate loss defined on the discrepancy between forward and backward costate rollouts, which are computationally cheap. We show that the costate dynamics, which can reflect state constraint violation, effectively enables the learning of differentiable values with large Lipschitz constants, without requiring manually supervised data as suggested by the hybrid PINN model. More importantly, we show that the close relationship between costates and policies makes the former critical in learning feedback control policies with generalizable safety performance.  

\end{abstract}


\vspace{-0.1in}
\section{Introduction}


We consider two-player general-sum differential games with deterministic dynamics, state constraints, and complete information. The Nash equilibrium values of such games satisfy a set of Hamilton-Jacobi-Isaacs (HJI) equations~\citep{viscosity, mitchell2005time,bressan2010noncooperative}. It is well known that approximating values and policies of differential games suffers from the following challenges: 
Firstly, alleviating the curse of dimensionality (CoD) through physics-informed neural networks (PINN) encounters convergence issues when value discontinuity (or differentiable value with large Lipschitz constant) is present due to state constraints~\citep{zhang2023value}: for system states and time starting from which a constraint cannot be satisfied eventually, the value becomes infinite (or large when constraints are treated as penalties).
Secondly, for policies to have good safety performance with respect to the state constraints, we need not only small approximation error for values but also for the gradients of values with respect to the states~\citep{yu2022gradient}.
Lastly, it is often necessary to learn generalizable values and policies across a parametric space of games, e.g., for the inference of game parameters when such information is private~\citep{cardaliaguet2012information}. In this paper, we focus on parameters that define large penalties that represent state constraints.

To address these challenges, we propose Pontryagin neural operator (PNO). PNO addresses the first and second challenges by introducing and minimizing costate losses defined on the discrepancy between forward and backward costate and value dynamics following Pontryagin's Maximum Principle (PMP)~\citep{mangasarian1966sufficient}: First, the costate dynamics that reflect state constraint violation~\citep{bokanowski2021relationship} can be readily computed by an ODE solver, and serves as a Lagrangian-frame self-supervised signal to facilitate the learning of highly nonlinear values, which cannot be achieved by the standard Eulerian-frame PINN. Second, the direct connection between costates and policies through the Hamiltonian makes costate-based learning more effective at converging to the ground-truth policies. For the last challenge on generalization to parametric PDEs, we extend our costate-based PINN from solving a single HJI equation to parametric HJIs using a DeepONet architecture~\citep{lu2021learning}, a neural operator that supports point-wise value gradient predictions needed for closed-loop control. 

\vspace{-0.12in}
\paragraph{Contributions.} 
\textbf{(1) Convergence without supervision:} The convergence issue in solving discontinuous HJI values via PINN has been investigated in~\cite{zhang2023approximating}, where the authors proposed to hybridize PINN using supervised equilibria data generated by solving games offline via PMP with sampled initial states. The major limitation of the hybrid method is in its assumption that knowledge already exists regarding where the informative equilibria should be sampled in the vast state space, e.g., collision and near-collision cases in vehicle interactions. In addition, generating supervisory data via solving PMP encounters its own numerical issues due to the existence of multiple equilibria and singular arcs. To the best of the authors' knowledge, this paper is among the first to address the PINN convergence issue \textit{without} using any supervised data or domain knowledge of corner cases. To do so, we sample costate trajectories based on a sampling distribution defined over the state space, and evolve this distribution along the training according to the costate loss landscape. Considering costate losses as PMP-driven constraints, the sampling distribution essentially captures the dual variables of the PINN. We conduct numerical studies using a two-vehicle game at an uncontrolled intersection.  
Our experimental results reveal that, with the same data complexity, PNO surpasses the hybrid neural operator (hybrid PINN + DeepONet) in safety performance across a range of parametric collision zones. 
\textbf{(2) Solving parametric HJI via value decomposition:} This paper is also among the first to study the efficacy of physics-informed neural operators in the context of parametric HJI equations. We show that the DeepONet architecture essentially learns a decomposition of the parametric value and identifies the major value basis functions. Empirical results on the same two-vehicle game show that the safety performance of the learned neural operator across a parametric set of collision zones closely aligns with the ground truth.
This investigation opens up new directions towards explainable data-driven models for policy learning: 
Can we translate basis value (or costate) functions into explainable sub-policies (e.g., temporal logic rules) that together comprise generalizable policies for parametric sets of games?

\vspace{-0.2in}
\section{Related Work}
\vspace{-0.05in}
\label{sec:related}

\paragraph{Differential games with state constraints.}
The existence of value for zero-sum differential games with Markovian payoffs and state constraints (and temporal logic specifications in general) have been derived~\citep{bettiol2006zero}. 
The results have been extended to general-sum games with non-temporal state constraints~\citep{zhang2018continuous}. 
To facilitate value approximation using level set methods, an epigraphical technique has been introduced to smooth discontinuous values when state constraints are present~\citep{gammoudi2023differential}.

\paragraph{HJI equations and physics-informed neural networks (PINN).}
HJI equations resulting from differential games are first-order nonlinear PDEs that suffer from CoD when solved by conventional level set methods~\citep{osher2004level, mitchell2005toolbox} or essentially nonoscillatory schemes~\citep{osher1991high}. 
Monte Carlo methods, such as variants of PINN~\citep{krishnapriyan2021characterizing}, have recently shown promise at solving high-dimensional PDEs, including HJ equations, provided that the solution is smooth~\citep{weinan2021algorithms}. 
PINN uses a trainable neural net to represent the solution space, and optimizes PDE-driven errors to approximate the solution. Such errors include the boundary residual~\citep{han2020convergence, han2018solving}, the PDE residual~\citep{jagtap2020adaptive,deepreach}, and supervisory data on the characteristic curves of the PDE~\citep{nakamura2021}.
Convergence and generalization of PINN have been analyzed under the assumption that both the solution and the network are Lipschitz continuous~\citep{han2020convergence,shin2020convergence,ito2021neural}. 
Recent studies have explored the effectiveness of PINN for solving PDEs with discontinuous solutions~\citep{jagtap2020adaptive} and solutions with large Kolmogorov n-width~\citep{mojgani2023kolmogorov}. However, solving PDEs with discontinuous solutions and only terminal boundaries (such as HJI equations with state constraints) is still an open challenge without prior structural knowledge about the value landscape~\citep{zhang2023value}.   

\vspace{-0.05in}
\paragraph{Neural operators.}
Neural operators emerged as a promising approach to universally approximating mappings between functions~\citep{kovachki2023neural}, with particularly successful applications to solving parametric PDEs~\citep{wang2021learning}. 
The architecture most related to this paper is DeepONet~\citep{lu2021learning}, which is composed of a branch net and a trunk net. The branch net extracts informative features of discretized input functions that define parameters of PDEs and the trunk net captures the basis functions that comprise parametric PDE solutions. DeepONet is one of the neural operator architectures that allow point-wise prediction, instead of predicting the entire function over its input domain (e.g., FNO~\citep{li2020fourier} and GNOT~\citep{hao2023gnot}). 
The extension of neural operators from supervised to PINN training has been studied for solving parametric physics equations defined on 2D and 3D state spaces~\citep{wang2021learning}. This paper examines the efficacy of the DeepONet architecture in physics-informed training of values with large Lipschitz constants defined on a 5D state-time space and a 2D parameter space. We compare two different neural operators empirically: the proposed Pontryagin neural operator and a hybrid one as a baseline.   

\vspace{-0.15in}
\section{Differential Games with Penalized State Constraints}
\vspace{-0.05in}
\label{sec:diff}

\paragraph{Notations and assumptions.}
\label{sec:notations}
Let the state dynamics of Player $i$ be $\dot{x}_i = f_i(x_i, u_i)$,
where $x_i\in\mathcal{X}_i\subseteq \mathbb{R}^{d_x}$ is the system state and $u_i\in\mathcal{U}_i\subseteq \mathbb{R}^{d_u}$ is the control input. The joint state space is $\mathcal{X} := \mathcal{X}_1 \times \mathcal{X}_2$. The instantaneous loss of Player $i$ is $l_i(\cdot, \cdot): \mathcal{X} \times \mathcal{U}_i \rightarrow \mathbb{R}$ and the terminal loss $g_i(\cdot): \mathcal{X}_i \rightarrow \mathbb{R}$. The fixed time horizon of the game is $[0, T]$. 
We use $\textbf{a}_i=(a_i, a_{-i})$ to concatenate elements $a_i$ from Player $i$ and $a_{-i}$ from the fellow player, and use $\textbf{a}=(a_1, a_2)$.
Denote $\alpha_i \in \mathcal{A}: \mathcal{X} \times [0,T] \rightarrow \mathcal{U}_i$ as Player $i$'s policy.
We use $x_s^{x_i,t,\alpha_i}$ as the state of Player $i$ at time $s$ if he follows policy $\alpha_i$, dynamics $f_i$, and starts at $(x_i,t)$. $\textbf{x}_s^{\textbf{x}_i,t,\boldsymbol{\alpha}_i}:=\left(x_s^{x_i,t,\alpha_i}, x_s^{x_{-i},t,\alpha_{-i}}\right)$.
Let $c_i(\cdot): \mathcal{X} \rightarrow \mathbb{R}$ be a state penalty derived from Player $i$'s state constraints, i.e., for any $\textbf{x}_i \in \mathcal{X}$, $c_i(\textbf{x}_i) = 0$ if
$\textbf{x}_i$ satisfies Player $i$'s state constraints, and otherwise $c_i(\textbf{x}_i)$ becomes a large positive number. In this paper, we consider $c_i(\cdot)$ to be differentiable but with a large Lipschitz constant.
The value of Player $i$ is denoted by $\vartheta_i(\cdot, \cdot): \mathcal{X} \times [0,T] \rightarrow \mathbb{R}$. 
We omit arguments to $f_i$, $l_i$, $g_i$, $c_i$, $\vartheta_i$ when possible, and decorate them with the superscript $\theta \in \Theta \subseteq \mathbb{R}^{d_{\theta}}$ when the corresponding functions are type-specific, where $\Theta$ is the type space. E.g., $c^{\theta}_i(\cdot,\cdot)$ is the state penalty of Player $i$ of type $\theta$.
Throughout the paper, we assume that $\mathcal{U}_i$ is compact and convex; $f_i$ and $c_i$ are Lipschitz continuous; $l_i$ and $g_i$ are Lipschitz continuous and bounded.

\vspace{-0.05in}
\paragraph{Value and HJI with state constraints.}
\label{sec:value}
Let $\boldsymbol{\alpha}^{\dagger}$ be a pair of equilibrium policies of the payoffs
\vspace{-0.05in}
\begin{equation}
 J_i(\textbf{x}_i, t, \boldsymbol{\alpha}_i) := \int_t^T \left(l_i\left(\textbf{x}_s^{\textbf{x}_i,t,\boldsymbol{\alpha}_i}, \alpha_i\left(\textbf{x}_s^{\textbf{x}_i,t,\boldsymbol{\alpha}_i}, s\right)\right) + c_i(\textbf{x}_s^{\textbf{x}_i,t,\boldsymbol{\alpha}_i}) \right)ds + g_i\left(\textbf{x}_T^{\textbf{x}_i,t,\boldsymbol{\alpha}_i}\right)
\label{eq:value_fun}
\vspace{-0.05in}
\end{equation}
for $i \in \{1,2\}$, so that
\vspace{-0.05in}
\begin{equation}
    J_i(\textbf{x}_i, t, \boldsymbol{\alpha}_i^{\dagger}) \leq J_i(\textbf{x}_i, t, (\alpha_i,\alpha_{-i}^{\dagger})), ~\forall \alpha_{i} \in \mathcal{A},~\forall i\in \{1,2\}.
\vspace{-0.05in}
\end{equation}
The value for Player $i$ is $\vartheta_i(\textbf{x}_i, t) = J_i(\textbf{x}_i, t, \boldsymbol{\alpha}^{\dagger}_i)$.
The HJI equations that govern the values, denoted by $L$, and the boundary conditions, by $D$, are the following~\cite{starr1969nonzero}:
\vspace{-0.05in}
\begin{equation}
\begin{aligned}
& L(\vartheta_i, \nabla_{\textbf{x}_i} \vartheta_i, \textbf{x}_i, t) := \nabla_t \vartheta_i + \max_{u \in \mathcal{U}_i} \left\{\nabla_{\textbf{x}_i} \vartheta_i ^T \textbf{f}_i - (l_i + c_i)\right\} = 0 \\
& D(\vartheta_i, \textbf{x}_i) := \vartheta_i(\textbf{x}_i, T) - g_i = 0, \quad \forall~ i = 1, 2. \\
\end{aligned}
\label{eq:hji}
\vspace{-0.05in}
\end{equation}
Therefore, Player $i$'s equilibrium policy can be derived as $\alpha_i^\dagger(\textbf{x}_i, t) = \argmax_{u \in \mathcal{U}_i} \{\nabla_{\textbf{x}_i} \vartheta_i^T \textbf{f}_i - (l_i + c_i) \}$ if Eq.~\eqref{eq:hji} can be solved for $(\vartheta_1, \vartheta_2)$~\citep{bressan2010noncooperative}. When needed, we denote by $\mathcal{L}^{\boldsymbol{\theta}}:=(L^{\boldsymbol{\theta}}, D^{\boldsymbol{\theta}})$ the HJI of a game parameterized by ${\boldsymbol{\theta}}$. 

\vspace{-0.05in}
\paragraph{Method of characteristics.}
The characteristic curves of $\mathcal{L}$ are open-loop equilibrium trajectories governed by PMP, which entails the following for initial state $(\bar{x}_1, \bar{x}_2) \in \mathcal{X}$ and $t \in [0,T]$:
\vspace{-0.05in}
\begin{equation}
\begin{aligned}
& \dot{x}_i = f_i, \quad x_i(t) = \bar{x}_i, \\
& \dot{\lambda}_i = - \nabla_{x_i} (\lambda_i^T \textbf{f}_i - (l_i+c_i)), \quad \lambda_i(T) = - \nabla_{x_i} g_i, \\
& u_i = \argmax_{u \in \mathcal{U}_i} ~\{\lambda_i^T\textbf{f}_i - (l_i+c_i)\},  \quad \forall~ i = 1, 2.
\end{aligned}
\label{eq:pmp}
\vspace{-0.05in}
\end{equation}
Here $\lambda_i = \nabla_{x_i}\vartheta_i$ is the costate of Player $i$. 
Solving Eq.~\eqref{eq:pmp} for a given initial states in $\mathcal{X}\times [0,T]$ is a boundary-value problem (BVP). Provided that convergence can be achieved, the solution to this BVP approximates $\boldsymbol{\vartheta}$ through the resultant characteristic trajectory and the boundary conditions $D$. \cite{zhang2023approximating} uses these characteristic trajectories as supervisory data, but assumes that informative initial states, i.e., those for which $c_i$ changes significantly along the trajectories, are known. PNO exploits the method of characteristics without requiring prior knowledge. See Sec.~\ref{sec:PNO}.

\vspace{-0.1in}
\section{Pontryagin Neural Operator}
\vspace{-0.05in}
\label{sec:PNO}
PNO is a neural operator $\hat{\vartheta}(\cdot,\cdot,\cdot): \mathcal{X} \times [0,T] \times \Theta^2 \rightarrow \mathbb{R}$ that maps ${\boldsymbol{\theta}} \in \Theta^2$ to  values of $\mathcal{L}^{\boldsymbol{\theta}}$. In the following, we first introduce the architecture of PNO and then explain the Pontryagin-mode physics-informed training, which is the key to its success.


\vspace{-0.05in}
\paragraph{Architecture.}
\label{sec: neural operator}
Following standard treatment in neural operators, we introduce an input function $a(\cdot,\cdot): \mathcal{X} \times \Theta^2 \rightarrow (0,1)$ to encode parametric settings of state constraints: $a(\textbf{x}, \boldsymbol{\theta}) = 1$ if $\textbf{x}$ violates the constraints according to $\boldsymbol{\theta}$, or otherwise $a(\textbf{x}, \boldsymbol{\theta}) = 0$. Let $X \in \mathbb{R}^{L \times d_x}$ be a lattice of $\mathcal{X}$, we denote by $a(X,\boldsymbol{\theta}) \in \{0,1\}^{L}$ the batch Boolean outputs at all $L$ lattice nodes.
Then $\hat{\vartheta}$ is defined as a linear combination of function bases: 
\begin{equation}
\hat{\vartheta}(\textbf{x}, t,\boldsymbol{\theta}) = \sum_{k=1}^q \underbrace{b_k(a(X, \boldsymbol{\theta}))}_{\rm branch} \underbrace{t_k(\textbf{x}, t)}_{\rm trunk},
\label{eq:DeepONet}
\end{equation}
where the branch network $\{b_k\}: \{0,1\}^L \rightarrow \mathbb{R}$ encodes the PDE parameters into function coefficients, and the trunk network $\{t_k\}: \mathbb{R}^{d_x} \times [0,T] \rightarrow \mathbb{R}$ encodes the input information into basis function values. 

\begin{figure}
\centering
\vspace{-15pt}
\includegraphics[width = 1\textwidth]{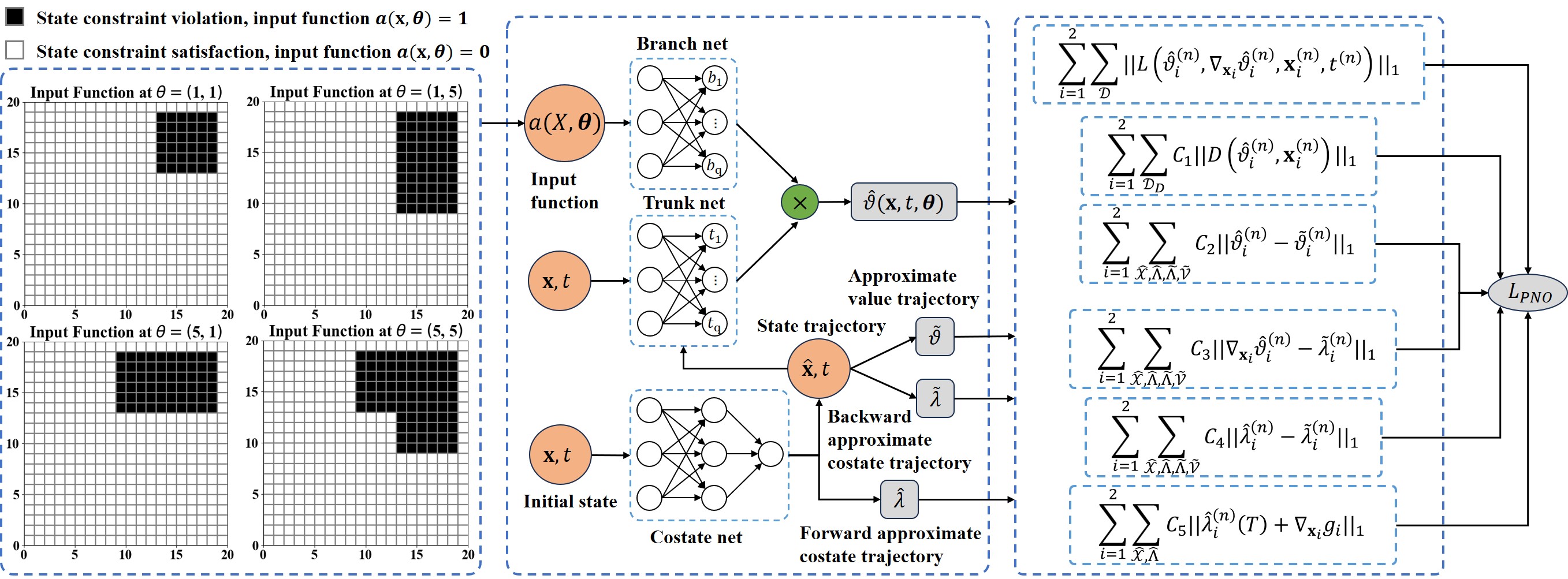}
\caption{Illustration of Pontryagin Neural Operator. PNO extends DeepONet~\citep{lu2021learning} to Pontryagin-mode PINN. The value $\hat{\vartheta}$ is decomposed into basis functions (trunk) and their HJI-parameter-dependent coefficients (branch). We use losses on costate predictions to  regularize the learning, which is key to the successful learning of values with large Lipschitz constants when state constraints are present. The baseline hybrid neural operator is similar to PNO, but without the learnable costate net. Instead, it solves BVPs as an overhead and use these fixed supervisory data for regularization.}
\label{fig: Pontryagin Onet}
\vspace{-20pt}
\end{figure}

\vspace{-0.05in}
\paragraph{Pontryagin-mode PINN.}
\label{sec: Pontryagain learning mode}

We introduce an additional costate network $\hat{\lambda}(\cdot,\cdot): \mathcal{X} \times [0,T] \rightarrow \mathcal{X}_*$, where $\mathcal{X}_*$ is the costate space. We can now evaluate the state, approximate costate, and approximate value trajectories starting from some sampled state and time $(\textbf{x},t)$ following their corresponding dynamics defined in Eq.~\eqref{eq:pmp}, first by sequentially maximizing the Hamiltonian using forward approximate $\hat{\lambda}$ to derive the state trajectory $\hat{\textbf{x}} \in \mathcal{X}^K$, and then using the transversality conditions of the approximate costate and value to derive approximate costate trajectory $\tilde{\lambda} \in \mathcal{X}_*^K$ and approximate value trajectory $\tilde{\vartheta} \in \mathbb{R}^K$ backward in time, where $K = \frac{T-t}{\Delta t}$ and $\Delta t$ is a small time interval. Given a batch of $N$ initial states, we collect forward approximate state trajectories $\hat{\mathcal{X}}:=\{\hat{\textbf{x}}^{(n)}\}_{n=1}^N$, forward and backward approximate costate trajectories, $\hat{\Lambda}:=\{\hat{\lambda}^{(n)}\}_{n=1}^N$ and $\tilde{\Lambda}:=\{\tilde{\lambda}^{(n)}\}_{n=1}^N$ respectively, and backward approximate value trajectories $\tilde{\mathcal{V}}:=\{\tilde{\vartheta}^{(n)}\}_{n=1}^N$. Lastly, we denote the conventional PINN dataset for formulating the PDE and boundary losses as $\mathcal{D}:=\{(\textbf{x},t)^{(n)} \in \mathcal{X} \times [0,T]\}_{n=1}^{N_L}$ and $\mathcal{D}_D:=\{\textbf{x}^{(n)} \in \mathcal{X}\}_{n=1}^{N_D}$. The PNO training loss with respect to value and costate functions of both players $(\hat{\boldsymbol{\vartheta}}, \hat{\boldsymbol{\lambda}})$ is then defined as:
\vspace{-0.05in}
\begin{equation}
\begin{aligned}
    L_{PNO}(\hat{\boldsymbol{\vartheta}}, \hat{\boldsymbol{\lambda}}) := 
    & \sum_{i=1}^2 \sum_{\mathcal{D}} \left\|L(\hat{\vartheta}_i^{(n)}, \nabla_{\textbf{x}_i} \hat{\vartheta}_i^{(n)}, \textbf{x}_i^{(n)}, t^{(n)}) \right\|_1 + \sum_{\mathcal{D}_D} C_1 \left\|D(\hat{\vartheta}_i^{(n)}, \textbf{x}_i^{(n)}) \right\|_1 \\
    & + \sum_{\hat{\mathcal{X}},\hat{\Lambda},\tilde{\Lambda},\tilde{\mathcal{V}} } C_2\left \|\hat{\vartheta}_i^{(n)} - \tilde\vartheta_i^{(n)} \right \|_1 + C_3 \left \|\nabla_{\textbf{x}_i}\hat{\vartheta}_i^{(n)} - \tilde{\lambda}_i^{(n)} \right\|_1 + C_4 \left\|\hat{\lambda}_i^{(n)} - \tilde{\lambda}_i^{(n)} \right\|_1 \\
    & + \sum_{\hat{\mathcal{X}},\hat{\Lambda}} C_5 \left \|\hat{\lambda}_i^{(n)}(T)  + \nabla_{\textbf{x}_i} g_i \right\|_1,
\end{aligned}
\label{eq:loss_PNO}
\vspace{-0.05in}
\end{equation}
where $\hat{\vartheta}_i^{(n)}$, $\tilde{\vartheta}_i^{(n)}$, $\hat{\lambda}_i^{(n)}$, $\tilde{\lambda}_i^{(n)}$ are abbreviations for $\hat{\vartheta}(\textbf{x}_i^{(n)}, t^{(n)}, \boldsymbol{\theta}_i^{(n)})$, $\tilde{\vartheta}_i(\textbf{x}_i^{(n)}, t^{(n)}, \boldsymbol{\theta}_i^{(n)})$, $\hat{\lambda}_i(\textbf{x}_i^{(n)}, t^{(n)})$,  $\tilde{\lambda}(\textbf{x}_i^{(n)}, t^{(n)})$ respectively. $C_i >0$ for $i=1,..,5$ are hyperparameters that balance the loss terms. We summarize the training of PNO in Alg.~\ref{alg: Pontryagin Onet}. Note that for the Pontryagin-mode losses to be differentiable with respect to $(\hat{\boldsymbol{\vartheta}}, \hat{\boldsymbol{\lambda}})$, we use costate net and ODE solver (RK45) to compute the forward and backward trajectories $(\hat{\mathcal{X}},\hat{\Lambda},\tilde{\Lambda},\tilde{\mathcal{V}})$. Line 9-12 in Alg.~\ref{alg: Pontryagin Onet} outline the computation.


\vspace{-0.05in}
\paragraph{Sampling strategy.} For standard PINN residual and boundary datasets $\mathcal{D}$ and $\mathcal{D}_D$, we use a curriculum learning scheme following \cite{deepreach, krishnapriyan2021characterizing}, where states are sampled from an expanding time window starting from the terminal time, to observe temporal causality of value functions. For Pontryagin-mode trajectories $(\hat{\mathcal{X}},\hat{\Lambda},\tilde{\Lambda},\tilde{\mathcal{V}})$, our experimental results show that an importance sampling strategy is critical. We start by sampling $N$ initial states uniformly in $\mathcal{X}$. Subsequently, we follow the evolutionary sampling algorithm~\citep{daw2022mitigating} to compute absolute values of the PDE residual $r(\textbf{x}_i^{(n)})=\|L(\hat\vartheta_i^{(n)}, \nabla_{\textbf{x}_i^{(n)}} \hat\vartheta_i^{(n)}, \textbf{x}_i^{(n)}, t^{(n)})\|_1$ for $n=1,...,N$ using Eq.~\eqref{eq:hji}\footnote{Our empirical studies showed that evolution based on costate losses achieves similar safety performance. Analytical understanding about the sampling-performance relation is yet to be established.}. We iteratively sample and remove samples with residuals lower than the average. 
The resultant batch represents initial states for which the current $\hat{\boldsymbol{\vartheta}}$ cannot generalize well. 
The sampled initial states gradually accumulate in the region with informative collision and near-collision knowledge during the training. Convergence of the evolutionary-based PINN has been proved in \cite{daw2022mitigating}.

\vspace{-0.1in}
\paragraph{Remarks.} 
PNO does not require offline data collection through solving BVPs and thus avoids typical convergence issues encountered in solving differential games with multiple local equilibria and slow convergence dynamics. 
While a full probably approximately correct (PAC) learning proof of PNO is not yet available, our results suggest that PNO achieves more data-efficient learning of values and policies than an existing hybrid method that takes advantage of informative samples that reveal structures of the value or costate landscapes. One hypothesis for its success is that PNO decomposes the value landscape into (terminal) boundaries and value dynamics along trajectories (a Lagrangian perspective), i.e., the hard-to-learn value with large Lipschitz constant can potentially be learned more effectively when each of its components, namely, the continuous boundary, continuous trajectories, and discontinuous value dynamics, are easy to learn given the others. We note that the improvement in data-efficiency of PINN through the method of characteristics has been discussed in \cite{mojgani2023kolmogorov} for 2D convection-diffusion and Burger's equations, but not yet in the context of high-dimensional HJ equations for optimal control or differential games.     

\vspace{-0.1in}
\section{Experiments and Results}
\vspace{-0.05in}
\label{sec: experiments}
To demonstrate the efficacy of PNO, we solve HJI equations for the interaction between two vehicles at an uncontrolled intersection. Each vehicle has two state variables (location $d_i$ and velocity $v_i$) and one parameter (player type $\theta$) that defines their perception of the collision zone. Hence we have a pair of 5D HJI equations defined in a 2D parameter space. We compare PNO with the hybrid method from \cite{zhang2023approximating} using both sample equilibrial trajectories solved from BVP (Eq.~\eqref{eq:pmp}) and the value landscape solved by a general-sum extension of the level set algorithm~\citep{bui2022optimizeddp}, which cannot be scaled to larger problems. Due to space limitation, we will focus on closed-loop safety performance as the performance metric. 
\textbf{Hardware.} Supervised equilibria data and closed-loop intersection simulations are computed on a workstation with 3.50GHz Xeon E5-1620 v4 and one GTX TITAN X with 12 GB memory. Hybrid and Pontryagin neural operators are trained on an A100 GPU with 40 GB memory.

\setlength{\algomargin}{1.5em}
\begin{algorithm2e}[h]
\caption{Pontryagin Neural Operator}
\label{alg: Pontryagin Onet}
\scriptsize
\LinesNumbered
\SetAlgoLined
\SetKwInOut{Input}{input}
\SetKwInOut{Output}{output}
\Input{$\mathcal{X}$, $\mathcal{D}$, $\mathcal{D}_D$, $T$, $pretrain\_iters$, $train\_iters$, $gradient\_steps$, learning rate $\alpha$, network parameters $\omega$}
\Output{$\hat{\vartheta}_{\omega}$, $\hat{\lambda}_{\omega}$}
initialize branch net $\{b_k\}$, trunk net $\{t_k\}$, and costate net $\hat{\lambda}$

$\mathcal{D}_D \leftarrow$ sample $\textbf{x} \in \mathcal{X}$ for $\{t_k\}$, and sample $\textbf{x} \in \mathcal{X}$ for $\hat{\lambda}$ at each pretrain iteration

pretrain $\{b_k\}$, $\{t_k\}$, and $\hat{\lambda}$ to satisfy value and costate boundary conditions via Eq.~\eqref{eq:hji} and Eq.~\eqref{eq:pmp} for $pretrain\_iters$ iterations

set $iter=0$, $num\_epoch=0$

\While{$iter \leq train\_iters$}{
$\mathcal{D}_D \leftarrow$ sample $\textbf{x} \in \mathcal{X}$ and $\mathcal{D} \leftarrow$ sample $\textbf{x} \in \mathcal{X}$ for $\{t_k\}$; set $\hat{\Lambda}=\emptyset$, $\tilde{\Lambda}=\emptyset$, $\tilde{\mathcal{V}}=\emptyset$, and $\hat{\mathcal{X}}=\emptyset$

\uIf{num\_epoch mod 10 == 0}{   
$\mathcal{D} \leftarrow$ sample $t \in [0, \frac{num\_epoch + 10}{train\_iters}T]$ for $\{t_k\}$ \tcc*[f]{Increase time window every 10 epochs}

sample $\textbf{x}(0) \in \mathcal{X}$ for $\hat{\lambda}$; $\hat{\mathcal{X}} \leftarrow$ compute state trajectories via solving Eq.~\eqref{eq:pmp} using $\hat{\lambda}$

$\hat{\Lambda} \leftarrow$ compute forward approximate costate trajectories using $\hat{\lambda}$ 

$\tilde{\Lambda} \leftarrow$ compute backward approximate costate trajectories as solutions to inverse value problems with terminal values $\hat{\mathcal{X}}(T)$ and $\hat{\Lambda}(T)$ via solving Eq.~\eqref{eq:pmp} using an ODE solver (RK45) 

$\tilde{\mathcal{V}} \leftarrow$ compute approximate value trajectories with $\hat{\mathcal{X}}$ and $\tilde{\Lambda}$ via solving Eq.~\eqref{eq:value_fun} and Eq.~\eqref{eq:pmp}
}

$num\_epoch \leftarrow num\_epoch+1$

compute $L_{PNO}$ via Eq.~\eqref{eq:loss_PNO}

$step=0$

\While{$step \leq gradient\_steps$}{
$\omega \leftarrow \omega - \alpha\nabla_{\omega}L_{PNO}$

$step \leftarrow step+1$}

$iter \leftarrow iter+1$
}
\end{algorithm2e}

\vspace{-0.05in}
\paragraph{Experiment setup.} 
We adopt settings of the uncontrolled intersection case studied in \cite{zhang2023approximating}: Each vehicle is represented by states $x_i := (d_i, v_i)$. Both vehicles follow dynamics: $\dot{d}_i = v_i$ and $\dot{v}_i = u_i$, where $u_i \in [-5, 10] m/s^2$ is the control input. Let $R$, $L$, and $W$ be the road length, vehicle length, and vehicle width, respectively. The instantaneous loss and state constraint are defined as: 
\vspace{-0.05in}
\begin{equation}
    {l}_i^{\theta}(\textbf{x}_i, u_i) = u_i^2, \quad c_i^{\theta}(\textbf{x}_i) =  b \sigma(d_i,\theta_i)\sigma(d_{-i},1),
\vspace{-0.05in}
\end{equation}
where 
\vspace{-0.05in}
\begin{equation}
    \sigma(d,\theta) = \left(1+\exp(-\gamma (d-R/2+\theta W/2))\right)^{-1}
    \left(1+\exp(\gamma (d-(R+W)/2-L))\right)^{-1}.
\end{equation}
Here shape parameter $\gamma = 5$, and $b = 10^4$ is sufficiently large to prevent collisions. Similar to treating collision as hard constraints, this setting causes the PINN convergence issue. $\theta \in \Theta := \{1, 2, 3, 4, 5\}$ denotes the aggressiveness level (safety preference) of a player. The terminal loss is defined to encourage players to pass the intersection and regain nominal speed:
\vspace{-0.05in}
\begin{equation}
g_i(x_i) = -\mu d_i(T) + (v_{i}(T)-\bar{v})^2,
\label{eq:case 1_terminal loss}
\vspace{-0.05in}
\end{equation}
where $\mu = 10^{-6}$, $\bar{v} = 18 m/s$, and $T = 3s$.

\begin{table}[t]
\vspace{-10pt}
\centering
\caption{Safety performance comparison among Hybrid neural operator and PNO, respectively for each parameter configuration in $\Theta^2$. Ground truth via BVP solver considers inevitable collisions.}
\vspace{-5pt}
\label{table: collision case}
\begin{adjustbox}{max width=\textwidth}
{\begin{tabular}{|c|c|c|c|c|c|c|c|c|c|c|c|c|c|c|c|} 
\hline  
Player Types & \multicolumn{15}{|c|}{Learning Method}\\
\hline  
Metrics & \multicolumn{15}{|c|}{Ground Truth \quad \textbar \quad Hybrid Neural Operator \quad \textbar \quad Pontryagin Neural Operator}\\
\hline  
& \multicolumn{3}{|c|}{(1, 1)} & \multicolumn{3}{|c|}{(1, 2)} & \multicolumn{3}{|c|}{(1, 3)} & \multicolumn{3}{|c|}{(1, 4)} & \multicolumn{3}{|c|}{(1, 5)}\\
\cline{2-16}
& 0.00\% & \textbf{0.17\%} & \textbf{0.17\%} & 2.67\% & 19.5\% & \textbf{6.50\%} & 4.83\% & 10.2\% & \textbf{9.17\%} & 9.33\% & 8.83\% & \textbf{8.00\%} & 9.00\% & \textbf{7.50\%} & 8.17\%\\
\cline{2-16}
& \multicolumn{3}{|c|}{(2, 1)} & \multicolumn{3}{|c|}{(2, 2)} & \multicolumn{3}{|c|}{(2, 3)} & \multicolumn{3}{|c|}{(2, 4)} & \multicolumn{3}{|c|}{(2, 5)}\\
\cline{2-16}
& 2.67\% & 19.5\% & \textbf{6.50\%} & 7.00\% & 24.2\% & \textbf{8.33\%} & 14.0\% & 23.8\% & \textbf{15.2\%} & 21.5\% & 26.5\% & \textbf{21.5\%} & 28.7\% & 30.2\% & \textbf{26.8\%}\\
\cline{2-16}
($\theta_1$, $\theta_2$) & \multicolumn{3}{|c|}{(3, 1)} & \multicolumn{3}{|c|}{(3, 2)} & \multicolumn{3}{|c|}{(3, 3)} & \multicolumn{3}{|c|}{(3, 4)} & \multicolumn{3}{|c|}{(3, 5)}\\
\cline{2-16}
Col.\% $\downarrow$ & 4.83\% & 10.2\% & \textbf{9.17\%} & 14.0\% & 23.8\% & \textbf{15.2\%} & 23.7\% & 26.5\% & \textbf{23.5\%} & 33.3\% & \textbf{32.3\%} & 33.2\% & 43.0\% & \textbf{41.0\%} & 41.3\%\\
\cline{2-16}
& \multicolumn{3}{|c|}{(4, 1)} & \multicolumn{3}{|c|}{(4, 2)} & \multicolumn{3}{|c|}{(4, 3)} & \multicolumn{3}{|c|}{(4, 4)} & \multicolumn{3}{|c|}{(4, 5)}\\
\cline{2-16}
& 9.33\% & 8.83\% & \textbf{8.00\%} & 21.5\% & 26.5\% & \textbf{21.5\%} & 33.3\% & \textbf{32.3\%} & 33.2\% & 42.3\% & \textbf{40.3\%} & 41.3\% & 54.3\% & \textbf{51.6\%} & 52.3\%\\
\cline{2-16}
& \multicolumn{3}{|c|}{(5, 1)} & \multicolumn{3}{|c|}{(5, 2)} & \multicolumn{3}{|c|}{(5, 3)} & \multicolumn{3}{|c|}{(5, 4)} & \multicolumn{3}{|c|}{(5, 5)}\\
\cline{2-16}
& 9.00\% & \textbf{7.50\%} & 8.17\% & 28.7\% & 30.2\% & \textbf{26.8\%} & 43.0\% & \textbf{41.0\%} & 41.3\% & 54.3\% & \textbf{51.6\%} & 52.3\% & 61.0\% & \textbf{59.7\%} & 60.2\%\\
\hline 
\end{tabular}} 
\end{adjustbox}
\vspace{-15pt}
\end{table}

\vspace{-0.05in}
\paragraph{Data.} 
To demonstrate the advantage of PNO, we perform a comparison in favor of the baseline (the hybrid neural operator): For the baseline, we generate 1k ground truth trajectories from initial states uniformly sampled in an informed state set $\mathcal{X}_{GT}:= [15, 20]m \times [18, 25]m/s$ by solving Eq.~\eqref{eq:pmp}. This set is identified to contain a significant amount of corner cases (e.g., $61\%$ of $\mathcal{X}_{GT}$ yield collisions when $\boldsymbol{\theta} = (5,5)$, see Tab.~\ref{table: collision case}) and will also be used for test. Each trajectory consists of 31 data points with a time interval of $0.1s$ for each of the players, resulting in a total of 62k data points. For PNO, the costate net explores the initial state within a larger state set $\mathcal{X}_{HJ}:= [15, 105]m \times [15, 32]m/s$ and generates 1k closed-loop trajectories including 62k data points. $\mathcal{X}_{HJ}$ is much less informed than $\mathcal{X}_{GT}$ (less collisions). To improve training efficiency, we eliminate data points generated by the costate net that fall beyond the state bounds and normalize the input data to $[-1, 1]$ for the trunk and costate net of both neural operators. Additionally, the trunk net uniformly samples 60k states in $\mathcal{X}_{HJ}:= [15, 105]m \times [15, 32]m/s$ for model refinement. Lastly, we train the model using four player-type configurations $(\theta_1, \theta_2)=\{(1, 1), (1, 5), (5, 1), (5, 5)\}$ and evaluate the model generalization performance within the type space $\Theta$. We reiterate that the total number of data points for the two methods are approximately the same.

\begin{table}[t]
\centering
\caption{Safety performance comparison among Hybrid neural operator and PNO, respectively for each parameter configuration in $\Theta^2$. Ground truth via BVP solver omits inevitable collisions.}
\vspace{-5pt}
\label{table: collision-free case}
\begin{adjustbox}{max width=\textwidth}
{\begin{tabular}{|c|c|c|c|c|c|c|c|c|c|c|c|c|c|c|c|} 
\hline  
Player Types & \multicolumn{15}{|c|}{Learning Method}\\
\hline  
Metrics & \multicolumn{15}{|c|}{Ground Truth \quad \textbar \quad Hybrid Neural Operator \quad \textbar \quad Pontryagin Neural Operator}\\
\hline  
& \multicolumn{3}{|c|}{(1, 1)} & \multicolumn{3}{|c|}{(1, 2)} & \multicolumn{3}{|c|}{(1, 3)} & \multicolumn{3}{|c|}{(1, 4)} & \multicolumn{3}{|c|}{(1, 5)}\\
\cline{2-16}
& 0.00\% & \textbf{0.17\%} & \textbf{0.17\%} & 0.00\% & 17.0\% & \textbf{4.33\%} & 0.00\% & 6.33\% & \textbf{5.67\%} & 0.00\% & 1.00\% & \textbf{0.00\%} & 0.00\% & \textbf{0.50\%} & 1.00\%\\
\cline{2-16}
& \multicolumn{3}{|c|}{(2, 1)} & \multicolumn{3}{|c|}{(2, 2)} & \multicolumn{3}{|c|}{(2, 3)} & \multicolumn{3}{|c|}{(2, 4)} & \multicolumn{3}{|c|}{(2, 5)}\\
\cline{2-16}
& 0.00\% & 17.0\% & \textbf{4.33\%} & 0.00\% & 18.8\% & \textbf{2.83\%} & 0.00\% & 11.3\% & \textbf{3.83\%} & 0.00\% & 7.17\% & \textbf{1.17\%} & 0.00\% & 4.50\% & \textbf{0.00\%}\\
\cline{2-16}
($\theta_1$, $\theta_2$) & \multicolumn{3}{|c|}{(3, 1)} & \multicolumn{3}{|c|}{(3, 2)} & \multicolumn{3}{|c|}{(3, 3)} & \multicolumn{3}{|c|}{(3, 4)} & \multicolumn{3}{|c|}{(3, 5)}\\
\cline{2-16}
Col.\% $\downarrow$ & 0.00\% & 6.33\% & \textbf{5.67\%} & 0.00\% & 11.3\% & \textbf{3.83\%} & 0.00\% & 5.33\% & \textbf{0.67\%} & 0.00\% & \textbf{0.33\%} & 0.50\% & 0.00\% & 0.83\% & \textbf{0.67\%}\\
\cline{2-16}
& \multicolumn{3}{|c|}{(4, 1)} & \multicolumn{3}{|c|}{(4, 2)} & \multicolumn{3}{|c|}{(4, 3)} & \multicolumn{3}{|c|}{(4, 4)} & \multicolumn{3}{|c|}{(4, 5)}\\
\cline{2-16}
& 0.00\% & 1.00\% & \textbf{0.00\%} & 0.00\% & 7.17\% & \textbf{1.17\%} & 0.00\% & \textbf{0.33\%} & 0.50\% & 0.00\% & \textbf{0.00\%} & \textbf{0.00\%} & 0.00\% & \textbf{0.33\%} & \textbf{0.33\%}\\
\cline{2-16} 
& \multicolumn{3}{|c|}{(5, 1)} & \multicolumn{3}{|c|}{(5, 2)} & \multicolumn{3}{|c|}{(5, 3)} & \multicolumn{3}{|c|}{(5, 4)} & \multicolumn{3}{|c|}{(5, 5)}\\
\cline{2-16}
& 0.00\% & \textbf{0.50\%} & 1.00\% & 0.00\% & 4.50\% & \textbf{0.00\%} & 0.00\% & 0.83\% & \textbf{0.67\%} & 0.00\% & \textbf{0.33\%} & \textbf{0.33\%} & 0.00\% & \textbf{0.00\%} & \textbf{0.00\%}\\
\hline 
\end{tabular}} 
\end{adjustbox}
\vspace{-15pt}
\end{table}

\begin{figure}[!ht]
\centering
\vspace{-10pt}
\includegraphics[width=0.85\linewidth]{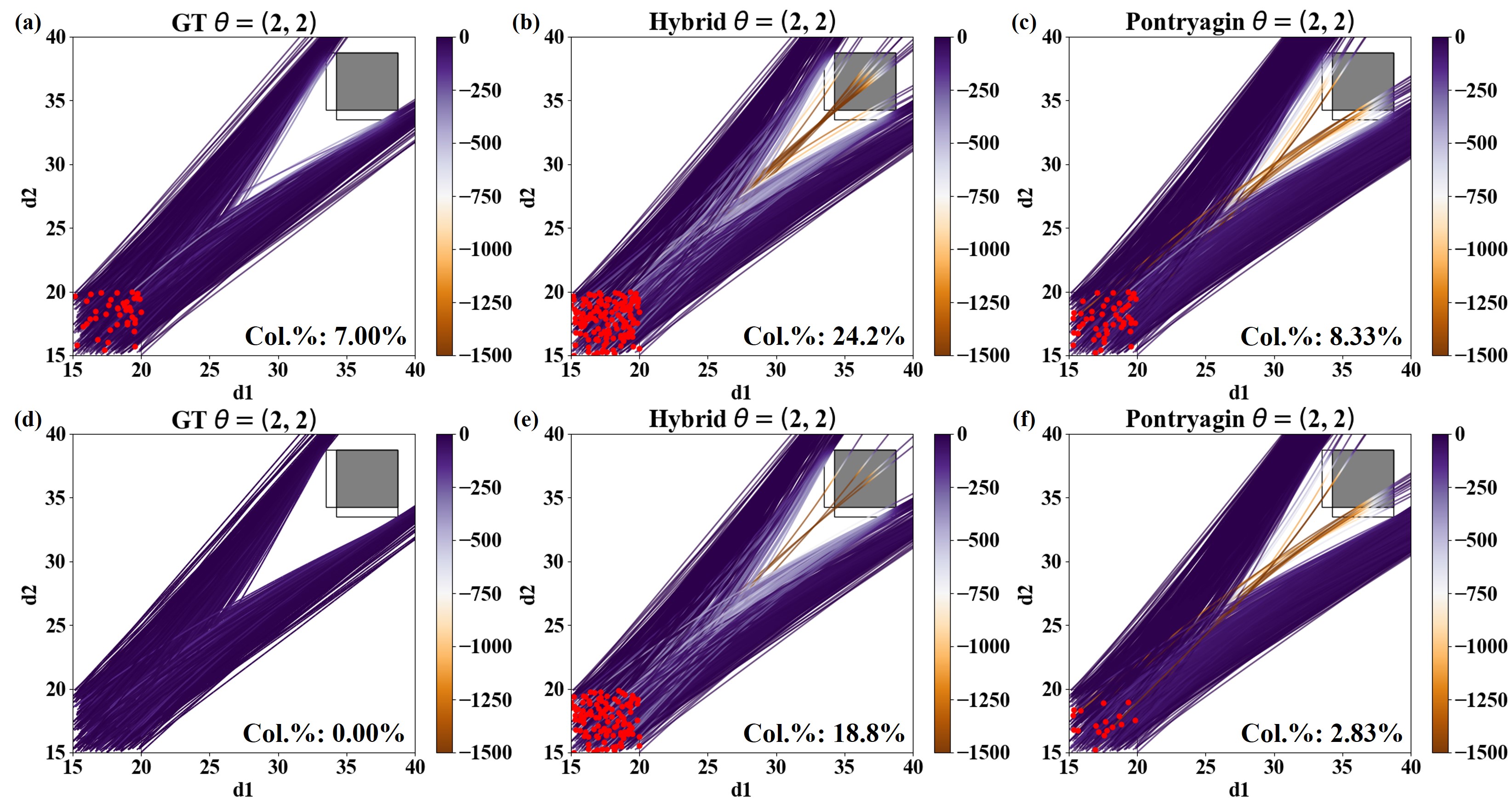}
\vspace{-10pt}
\caption{(a, d): Ground truth trajectories (projected to $d_1$-$d_2$) w and w/o inevitable collisions, respectively. (b-c), (e-f): Trajectories generated using hybrid and PNO w and w/o inevitable collisions, respectively. Color: Actual values of Player 1. Red dots: Initial states that yield collisions. Solid gray box: collision area for player type $\boldsymbol{\theta}=(1,1)$. Players with larger type parameters (e.g., $\boldsymbol{\theta}=(2,2)$) have correspondingly larger collision zones, as indicated by white box.}
\label{fig: safety performance}
\vspace{-15pt}
\end{figure}

\vspace{-0.05in}
\paragraph{Training.} 
The neural operators utilize fully-connected networks with 3 hidden layers, each comprising 64 neurons with \texttt{tanh} activation. The hybrid neural operator uses the Adam optimizer with a fixed learning rate of $2 \times 10^{-5}$ for pre-training the network over 100k iterations using supervised data. It then combines the supervised data with states sampled from an expanding time window, starting from the terminal time, to refine the model for an additional 200k iterations. PNO uses the Adam optimizer with an adaptive learning rate, starting from $2 \times 10^{-5}$, and is initially trained to satisfy the boundary conditions over 50k iterations. Subsequently, the network undergoes refinement through 3k gradient steps per epoch, encompassing a total of 10 epochs for every 10 training iterations. The costate net resamples the initial state to generate closed-loop trajectories every 10 epochs, and these data points are used to train the model. The overall network refinement process spans 300 training iterations. Both PNO and hybrid incorporate adaptive activation functions~\citep{jagtap2020adaptive}. In terms of wall-clock computational cost, the hybrid neural operator requires approximately 37 hours, including 5.5 hours for data acquisition and 31.5 hours for model training. PNO requires 27 hours for model training. For comparison, solving all 25 individual HJI PDEs in $\Theta^2$ with an acceptable meshgrid resolution using an efficient level set package~\citep{bui2022optimizeddp} would require $>50$ hours. The resultant value approximation does not guarantee safety performance since value gradient approximation error is not controlled. 

\begin{figure}
\centering
\vspace{-15pt}
\includegraphics[width = 0.8\textwidth]{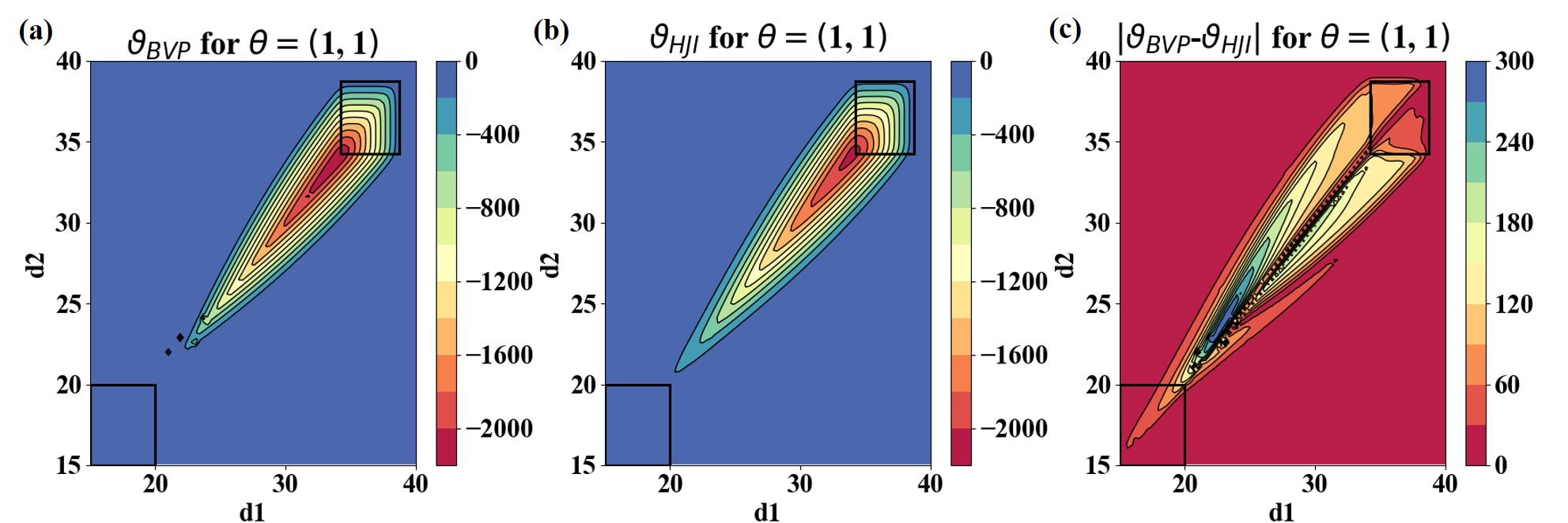}
\vspace{-10pt}
\caption{(a, b) Value contour of BVP solutions and HJI equations projected respectively to $(d_1,d_2)$ with fixed $v_{1,2} = 18m/s$ and $t=0$ for $\boldsymbol{\theta}=(1,1)$. (c) Difference $|\vartheta_{BVP}-\vartheta_{HJI}|$.}
\label{fig: bvp_hji}
\vspace{-20pt}
\end{figure}

\vspace{-0.09in}
\paragraph{Safety performance.}
We use the neural operators to compute $\nabla_{\textbf{x}_i}\hat\vartheta_i, \forall i=1,2$ as closed-loop control on test cases uniformly sampled in $\mathcal{X}_{GT}$ and report their safety performance in percentage of collisions, where collisions are defined by $\boldsymbol{\theta}$. For better transparency, the comparison uses two sets of ground truth trajectories: The first contains 600 trajectories for each parameter configuration in $\Theta^2$; the second uses the same setting but only contains trajectories without physical collisions (for initial states from which collision cannot be avoided according to $\boldsymbol{\theta} = (1,1)$). Tab.~\ref{table: collision case} and Tab.~\ref{table: collision-free case} summarize safety performance comparisons with and without inevitable collisions, generalizing to testing cases across $\Theta^2$. The results show that PNO outperforms the hybrid neural operator in most cases. More importantly, it achieves this without relying on domain knowledge (informative trajectories in $\mathcal{X}_{GT}$). Trajectories from test player types are visualized in Fig.~\ref{fig: safety performance}. It should be noted that in some cases neural operators perform better than the ground truth. This is because a multi-start BVP solver solves the ground truth trajectories with heuristic initial guesses. We also note that while BVP solutions are open-loop, their corresponding values are consistent with the HJI equations for this case study (see comparison between BVP and HJI value contours at $t=0$ in Fig.~\ref{fig: bvp_hji}).

\begin{figure}
\centering
\vspace{-15pt}
\includegraphics[width = 1\textwidth]{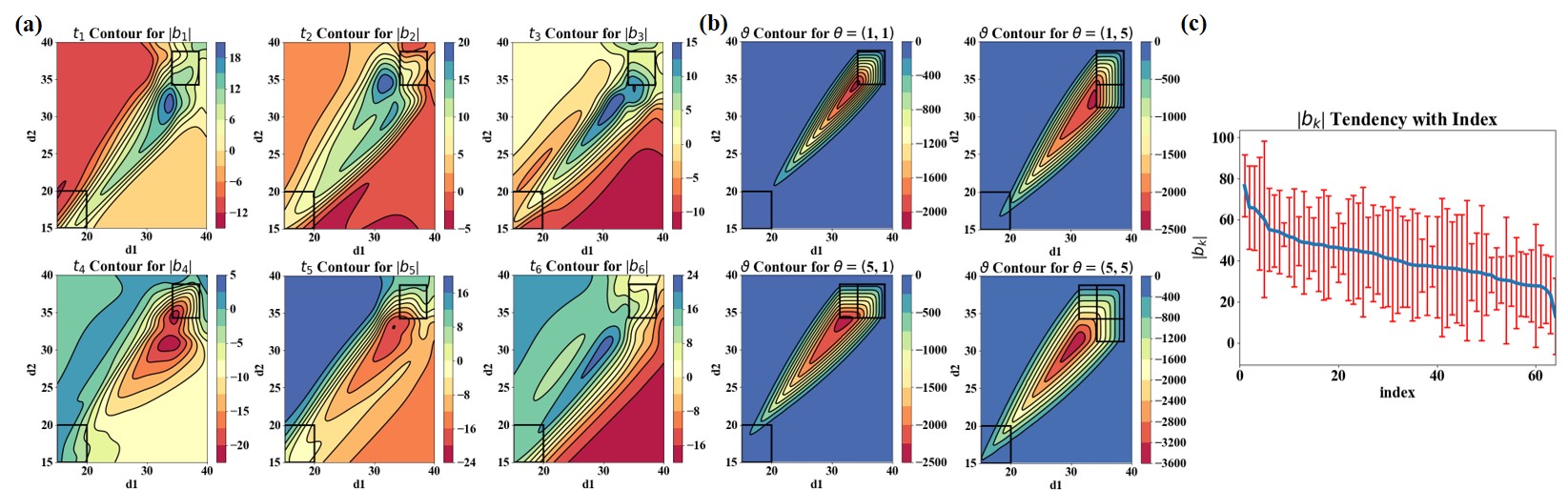}
\vspace{-20pt}
\caption{(a) Basis value functions ($t_k$) corresponding to the largest mean $b_k$. Visualized for $(d_1,d_2)$ with fixed $v_{1,2} = 18m/s$ and $t=0$. (b) Ground truth values for sampled $\boldsymbol{\theta}$. (c) Sorted sample mean and std of coefficients $|b_k|$ over $\Theta^2$.}
\label{fig: deeponet}
\vspace{-20pt}
\end{figure}

\vspace{-0.09in}
\paragraph{Structure of parametric value.} Since branch and trunk nets learn a decomposition of the value function in the parameter space, Fig.~\ref{fig: deeponet} investigates the learned value structure through visualization of (1)  the major basis value functions associated with the 6 leading means of absolute coefficients ($b_k$ in Eq.~\eqref{eq:DeepONet}) over $\Theta^2$ (Fig.~\ref{fig: deeponet}a), (2) the ground truth values solved using level set for $\boldsymbol{\theta}=(1,1), ~(1,5), ~(5,1), ~(5,5)$ (Fig.~\ref{fig: deeponet}b), and (3) the sorted mean and standard deviation of absolute coefficients over $\Theta^2$ along the output dimension of branch net, where the dominant basis value functions correspond to larger $|b_k|$ (Fig.~\ref{fig: deeponet}c). Further regularization on the branch net to promote basis sparsity and extension of existing generalization results to neural operators are worth investigating in future studies.   

\begin{figure}[!ht]
\centering
\vspace{-5pt}
\includegraphics[width=0.85\linewidth]{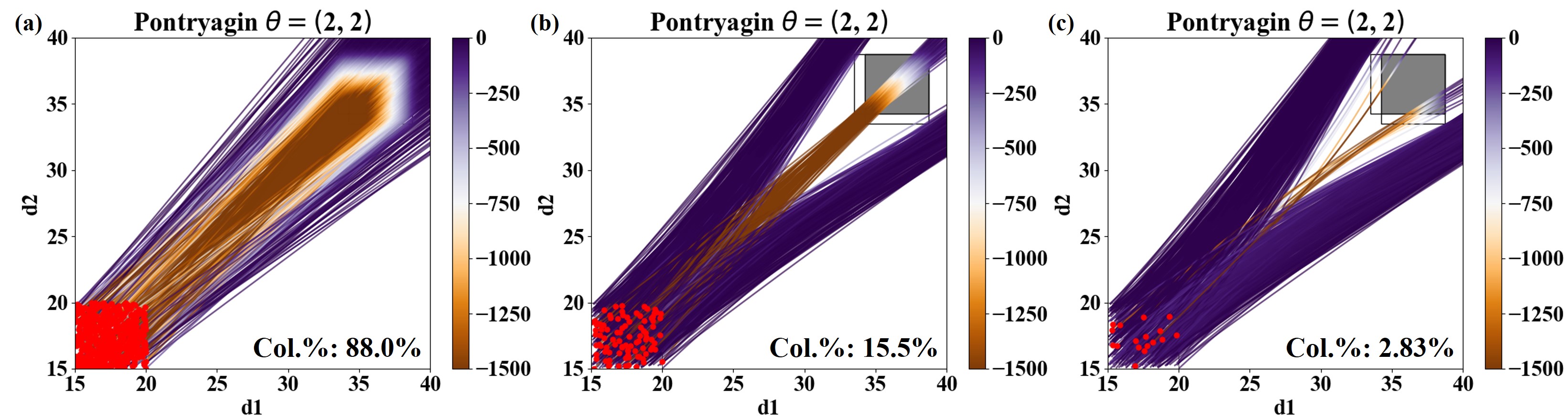}
\vspace{-5pt}
\caption{Closed-loop trajectories generated using PNO with (a) \texttt{relu}, (b) \texttt{sin} and (c) \texttt{tanh} activation functions without inevitable collisions.}
\label{fig: activation}
\vspace{-5pt}
\end{figure}

\vspace{-0.09in}
\paragraph{Ablation studies.} 
Due to the reported importance of activation choice in PINN, we conduct ablation studies to understand the influence of activation on safety performance. Tab.~\ref{table: activation function collision-free case} summarizes the safety performance using different activation functions for comparisons w/ and w/o inevitable collisions. The comparison of closed-loop trajectories is visualized in Fig.~\ref{fig: activation}. The results confirm that the choice of activation affects safety performance: \texttt{tanh} outperforms \texttt{sin} and \texttt{relu}.

\begin{table}[!ht]
\vspace{-20pt}
\centering
\caption{Safety comparison among activation functions using PNO w/o inevitable collisions for all configurations in $\Theta^2$}
\vspace{-0.10in}
\label{table: activation function collision-free case}
\begin{adjustbox}{max width=\textwidth}
{\begin{tabular}{|c|c|c|c|c|c|c|c|c|c|c|c|c|c|c|c|} 
\hline  
Player Types & \multicolumn{15}{|c|}{Activation Function}\\
\hline  
Metrics & \multicolumn{15}{|c|}{\texttt{relu} \quad \textbar \quad \texttt{sin} \quad \textbar \quad \texttt{tanh}}\\
\hline  
& \multicolumn{3}{|c|}{(1, 1)} & \multicolumn{3}{|c|}{(1, 2)} & \multicolumn{3}{|c|}{(1, 3)} & \multicolumn{3}{|c|}{(1, 4)} & \multicolumn{3}{|c|}{(1, 5)}\\
\cline{2-16}
& 81.3\% & 1.83\% & \textbf{0.17\%} & 84.8\% & \textbf{4.33\%} & \textbf{4.33\%} & 86.2\% & 12.5\% & \textbf{5.67\%} & 86.8\% & 5.33\% & \textbf{0.00\%} & 87.8\% & 3.50\% & \textbf{1.00\%}\\
\cline{2-16}
& \multicolumn{3}{|c|}{(2, 1)} & \multicolumn{3}{|c|}{(2, 2)} & \multicolumn{3}{|c|}{(2, 3)} & \multicolumn{3}{|c|}{(2, 4)} & \multicolumn{3}{|c|}{(2, 5)}\\
\cline{2-16}
& 84.8\% & \textbf{4.33\%} & \textbf{4.33\%} & 88.0\% & 15.5\% & \textbf{2.83\%} & 88.8\% & 12.0\% & \textbf{3.83\%} & 90.3\% & 4.33\% & \textbf{1.17\%} & 89.8\% & 0.67\% & \textbf{0.00\%}\\
\cline{2-16}
($\theta_1$, $\theta_2$) & \multicolumn{3}{|c|}{(3, 1)} & \multicolumn{3}{|c|}{(3, 2)} & \multicolumn{3}{|c|}{(3, 3)} & \multicolumn{3}{|c|}{(3, 4)} & \multicolumn{3}{|c|}{(3, 5)}\\
\cline{2-16}
Col.\% $\downarrow$ & 86.2\% & 12.5\% & \textbf{5.67\%} & 88.8\% & 12.0\% & \textbf{3.83\%} & 92.5\% & 4.17\% & \textbf{0.67\%} & 91.5\% & 0.83\% & \textbf{0.50\%} & 93.7\% & \textbf{0.67\%} & \textbf{0.67\%}\\
\cline{2-16}
& \multicolumn{3}{|c|}{(4, 1)} & \multicolumn{3}{|c|}{(4, 2)} & \multicolumn{3}{|c|}{(4, 3)} & \multicolumn{3}{|c|}{(4, 4)} & \multicolumn{3}{|c|}{(4, 5)}\\
\cline{2-16}
& 86.8\% & 5.33\% & \textbf{0.00\%} & 90.3\% & 4.33\% & \textbf{1.17\%} & 91.5\% & 0.83\% & \textbf{0.50\%} & 94.8\% & 0.33\% & \textbf{0.00\%} & 95.3\% & \textbf{0.17\%} & 0.33\%\\
\cline{2-16}
& \multicolumn{3}{|c|}{(5, 1)} & \multicolumn{3}{|c|}{(5, 2)} & \multicolumn{3}{|c|}{(5, 3)} & \multicolumn{3}{|c|}{(5, 4)} & \multicolumn{3}{|c|}{(5, 5)}\\
\cline{2-16}
& 87.8\% & 3.50\% & \textbf{1.00\%} & 89.8\% & 0.67\% & \textbf{0.00\%} & 93.7\% & \textbf{0.67\%} & \textbf{0.67\%} & 95.3\% & \textbf{0.17\%} & 0.33\% & 95.8\% & \textbf{0.00\%} & \textbf{0.00\%}\\
\hline 
\end{tabular}} 
\end{adjustbox}
\vspace{-10pt}
\end{table}

\vspace{-0.20in}
\section{Conclusion}
\vspace{-0.05in}
We introduced PNO to approximate parametric value functions with large Lipschitz constants as solutions to two-player general-sum differential games with state constraints. Compared with the existing PINN solution that requires informative supervisory data, PNO is fully self-supervised and free of prior knowledge. For this reason, PNO is also free of convergence issues related to the existence of multiple equilibria or singular arcs in solving BVPs. In the intersection case study, our numerical results demonstrated that PNO yields superior safety performance compared to the hybrid neural operator, achieving this with a lower computational cost. Compared with value iteration which uses Bellman equation in the Lagrangian frame (i.e., samples along trajectories), PNO jointly uses Bellman equation in the Eulerian frame (i.e., PINN loss) and the state gradient of Bellman in the Lagrangian frame (i.e., costate losses). The connection between PNO and value iteration in terms of their sampling complexity should be further explored. Code is available on the \href{https://github.com/dayanbatuofu/PNO}{GitHub}.

\acks{This work was in part supported by NSF CMMI-1925403 and NSF CNS-2101052. The views and conclusions contained in this document are those of the authors and should not be interpreted as representing the official policies, either expressed or implied, of the National Science Foundation or the U.S. Government.}

\bibliography{l4dc2024}

\begin{thebibliography}{33}
\providecommand{\natexlab}[1]{#1}
\providecommand{\url}[1]{\texttt{#1}}
\expandafter\ifx\csname urlstyle\endcsname\relax
  \providecommand{\doi}[1]{doi: #1}\else
  \providecommand{\doi}{doi: \begingroup \urlstyle{rm}\Url}\fi

\bibitem[Bansal and Tomlin(2021)]{deepreach}
Somil Bansal and Claire~J Tomlin.
\newblock {DeepReach}: A deep learning approach to high-dimensional
  reachability.
\newblock In \emph{2021 IEEE International Conference on Robotics and
  Automation (ICRA)}, pages 1817--1824. IEEE, 2021.

\bibitem[Bettiol et~al.(2006)Bettiol, Cardaliaguet, and
  Quincampoix]{bettiol2006zero}
Piernicola Bettiol, Pierre Cardaliaguet, and Marc Quincampoix.
\newblock {Zero-sum state constrained differential games: existence of value
  for Bolza problem}.
\newblock \emph{International Journal of Game Theory}, 34:\penalty0 495--527,
  2006.

\bibitem[Bokanowski et~al.(2021)Bokanowski, D{\'e}silles, and
  Zidani]{bokanowski2021relationship}
Olivier Bokanowski, Anya D{\'e}silles, and Hasnaa Zidani.
\newblock Relationship between maximum principle and dynamic programming in
  presence of intermediate and final state constraints.
\newblock \emph{ESAIM: Control, Optimisation and Calculus of Variations},
  27:\penalty0 91, 2021.

\bibitem[Bressan(2010)]{bressan2010noncooperative}
Alberto Bressan.
\newblock Noncooperative differential games. a tutorial.
\newblock \emph{Department of Mathematics, Penn State University}, 81, 2010.

\bibitem[Bui et~al.(2022)Bui, Giovanis, Chen, and
  Shriraman]{bui2022optimizeddp}
Minh Bui, George Giovanis, Mo~Chen, and Arrvindh Shriraman.
\newblock Optimizeddp: An efficient, user-friendly library for optimal control
  and dynamic programming.
\newblock \emph{arXiv preprint arXiv:2204.05520}, 2022.

\bibitem[Cardaliaguet(2012)]{cardaliaguet2012information}
Pierre Cardaliaguet.
\newblock Information issues in differential game theory.
\newblock In \emph{ESAIM: Proceedings}, volume~35, pages 1--13. EDP Sciences,
  2012.

\bibitem[Crandall and Lions(1983)]{viscosity}
Michael~G Crandall and Pierre-Louis Lions.
\newblock {Viscosity solutions of Hamilton-Jacobi equations}.
\newblock \emph{Transactions of the American mathematical society},
  277\penalty0 (1):\penalty0 1--42, 1983.

\bibitem[Daw et~al.(2022)Daw, Bu, Wang, Perdikaris, and
  Karpatne]{daw2022mitigating}
Arka Daw, Jie Bu, Sifan Wang, Paris Perdikaris, and Anuj Karpatne.
\newblock Mitigating propagation failures in pinns using evolutionary sampling.
\newblock 2022.

\bibitem[E et~al.(2021)E, Han, and Jentzen]{weinan2021algorithms}
Weinan E, Jiequn Han, and Arnulf Jentzen.
\newblock {Algorithms for solving high dimensional PDEs: from nonlinear Monte
  Carlo to machine learning}.
\newblock \emph{Nonlinearity}, 35\penalty0 (1):\penalty0 278, 2021.

\bibitem[Gammoudi and Zidani(2023)]{gammoudi2023differential}
Nidhal Gammoudi and Hasnaa Zidani.
\newblock A differential game control problem with state constraints.
\newblock \emph{Mathematical Control and Related Fields}, 13\penalty0
  (2):\penalty0 554--582, 2023.

\bibitem[Han and Long(2020)]{han2020convergence}
Jiequn Han and Jihao Long.
\newblock Convergence of the deep bsde method for coupled fbsdes.
\newblock \emph{Probability, Uncertainty and Quantitative Risk}, 5\penalty0
  (1):\penalty0 1--33, 2020.

\bibitem[Han et~al.(2018)Han, Jentzen, and E]{han2018solving}
Jiequn Han, Arnulf Jentzen, and Weinan E.
\newblock Solving high-dimensional partial differential equations using deep
  learning.
\newblock \emph{Proceedings of the National Academy of Sciences}, 115\penalty0
  (34):\penalty0 8505--8510, 2018.

\bibitem[Hao et~al.(2023)Hao, Wang, Su, Ying, Dong, Liu, Cheng, Song, and
  Zhu]{hao2023gnot}
Zhongkai Hao, Zhengyi Wang, Hang Su, Chengyang Ying, Yinpeng Dong, Songming
  Liu, Ze~Cheng, Jian Song, and Jun Zhu.
\newblock {GNOT: A general neural operator transformer for operator learning}.
\newblock In \emph{International Conference on Machine Learning}, pages
  12556--12569. PMLR, 2023.

\bibitem[Ito et~al.(2021)Ito, Reisinger, and Zhang]{ito2021neural}
Kazufumi Ito, Christoph Reisinger, and Yufei Zhang.
\newblock {A neural network-based policy iteration algorithm with global
  $H^{2}$-superlinear convergence for stochastic games on domains}.
\newblock \emph{Foundations of Computational Mathematics}, 21\penalty0
  (2):\penalty0 331--374, 2021.

\bibitem[Jagtap et~al.(2020)Jagtap, Kawaguchi, and
  Karniadakis]{jagtap2020adaptive}
Ameya~D Jagtap, Kenji Kawaguchi, and George~Em Karniadakis.
\newblock Adaptive activation functions accelerate convergence in deep and
  physics-informed neural networks.
\newblock \emph{Journal of Computational Physics}, 404:\penalty0 109136, 2020.

\bibitem[Kovachki et~al.(2023)Kovachki, Li, Liu, Azizzadenesheli, Bhattacharya,
  Stuart, and Anandkumar]{kovachki2023neural}
Nikola Kovachki, Zongyi Li, Burigede Liu, Kamyar Azizzadenesheli, Kaushik
  Bhattacharya, Andrew Stuart, and Anima Anandkumar.
\newblock Neural operator: Learning maps between function spaces with
  applications to pdes.
\newblock \emph{Journal of Machine Learning Research}, 24\penalty0
  (89):\penalty0 1--97, 2023.

\bibitem[Krishnapriyan et~al.(2021)Krishnapriyan, Gholami, Zhe, Kirby, and
  Mahoney]{krishnapriyan2021characterizing}
Aditi Krishnapriyan, Amir Gholami, Shandian Zhe, Robert Kirby, and Michael~W
  Mahoney.
\newblock Characterizing possible failure modes in physics-informed neural
  networks.
\newblock \emph{Advances in Neural Information Processing Systems},
  34:\penalty0 26548--26560, 2021.

\bibitem[Li et~al.(2020)Li, Kovachki, Azizzadenesheli, Liu, Bhattacharya,
  Stuart, and Anandkumar]{li2020fourier}
Zongyi Li, Nikola Kovachki, Kamyar Azizzadenesheli, Burigede Liu, Kaushik
  Bhattacharya, Andrew Stuart, and Anima Anandkumar.
\newblock Fourier neural operator for parametric partial differential
  equations.
\newblock \emph{arXiv preprint arXiv:2010.08895}, 2020.

\bibitem[Lu et~al.(2021)Lu, Jin, Pang, Zhang, and Karniadakis]{lu2021learning}
Lu~Lu, Pengzhan Jin, Guofei Pang, Zhongqiang Zhang, and George~Em Karniadakis.
\newblock {Learning nonlinear operators via DeepONet based on the universal
  approximation theorem of operators}.
\newblock \emph{Nature machine intelligence}, 3\penalty0 (3):\penalty0
  218--229, 2021.

\bibitem[Mangasarian(1966)]{mangasarian1966sufficient}
Olvi~L Mangasarian.
\newblock Sufficient conditions for the optimal control of nonlinear systems.
\newblock \emph{SIAM Journal on control}, 4\penalty0 (1):\penalty0 139--152,
  1966.

\bibitem[Mitchell and Templeton(2005)]{mitchell2005toolbox}
Ian~M Mitchell and Jeremy~A Templeton.
\newblock {A toolbox of Hamilton-Jacobi solvers for analysis of
  nondeterministic continuous and hybrid systems}.
\newblock In \emph{Hybrid Systems: Computation and Control: 8th International
  Workshop, HSCC 2005, Zurich, Switzerland, March 9-11, 2005. Proceedings 8},
  pages 480--494. Springer, 2005.

\bibitem[Mitchell et~al.(2005)Mitchell, Bayen, and Tomlin]{mitchell2005time}
Ian~M Mitchell, Alexandre~M Bayen, and Claire~J Tomlin.
\newblock A time-dependent hamilton-jacobi formulation of reachable sets for
  continuous dynamic games.
\newblock \emph{IEEE Transactions on automatic control}, 50\penalty0
  (7):\penalty0 947--957, 2005.

\bibitem[Mojgani et~al.(2023)Mojgani, Balajewicz, and
  Hassanzadeh]{mojgani2023kolmogorov}
Rambod Mojgani, Maciej Balajewicz, and Pedram Hassanzadeh.
\newblock {Kolmogorov n--width and Lagrangian physics-informed neural networks:
  a causality-conforming manifold for convection-dominated PDEs}.
\newblock \emph{Computer Methods in Applied Mechanics and Engineering},
  404:\penalty0 115810, 2023.

\bibitem[Nakamura-Zimmerer et~al.(2021)Nakamura-Zimmerer, Gong, and
  Kang]{nakamura2021}
Tenavi Nakamura-Zimmerer, Qi~Gong, and Wei Kang.
\newblock {Adaptive deep learning for high-dimensional
  Hamilton--Jacobi--Bellman equations}.
\newblock \emph{SIAM Journal on Scientific Computing}, 43\penalty0
  (2):\penalty0 A1221--A1247, 2021.

\bibitem[Osher and Shu(1991)]{osher1991high}
Stanley Osher and Chi-Wang Shu.
\newblock {High-order essentially nonoscillatory schemes for Hamilton--Jacobi
  equations}.
\newblock \emph{SIAM Journal on numerical analysis}, 28\penalty0 (4):\penalty0
  907--922, 1991.

\bibitem[Osher et~al.(2004)Osher, Fedkiw, and Piechor]{osher2004level}
Stanley Osher, Ronald Fedkiw, and K~Piechor.
\newblock Level set methods and dynamic implicit surfaces.
\newblock \emph{Appl. Mech. Rev.}, 57\penalty0 (3):\penalty0 B15--B15, 2004.

\bibitem[Shin et~al.(2020)Shin, Darbon, and Karniadakis]{shin2020convergence}
Yeonjong Shin, Jerome Darbon, and George~Em Karniadakis.
\newblock On the convergence of physics informed neural networks for linear
  second-order elliptic and parabolic type pdes.
\newblock \emph{arXiv preprint arXiv:2004.01806}, 2020.

\bibitem[Starr and Ho(1969)]{starr1969nonzero}
Alan~Wilbor Starr and Yu-Chi Ho.
\newblock Nonzero-sum differential games.
\newblock \emph{Journal of optimization theory and applications}, 3\penalty0
  (3):\penalty0 184--206, 1969.

\bibitem[Wang et~al.(2021)Wang, Wang, and Perdikaris]{wang2021learning}
Sifan Wang, Hanwen Wang, and Paris Perdikaris.
\newblock {Learning the solution operator of parametric partial differential
  equations with physics-informed DeepONets}.
\newblock \emph{Science advances}, 7\penalty0 (40):\penalty0 eabi8605, 2021.

\bibitem[Yu et~al.(2022)Yu, Lu, Meng, and Karniadakis]{yu2022gradient}
Jeremy Yu, Lu~Lu, Xuhui Meng, and George~Em Karniadakis.
\newblock {Gradient-enhanced physics-informed neural networks for forward and
  inverse PDE problems}.
\newblock \emph{Computer Methods in Applied Mechanics and Engineering},
  393:\penalty0 114823, 2022.

\bibitem[Zhang et~al.(2023{\natexlab{a}})Zhang, Ghimire, Zhang, Xu, and
  Ren]{zhang2023approximating}
Lei Zhang, Mukesh Ghimire, Wenlong Zhang, Zhe Xu, and Yi~Ren.
\newblock {Approximating discontinuous Nash equilibrial values of two-player
  general-Sum differential games}.
\newblock In \emph{2023 IEEE International Conference on Robotics and
  Automation (ICRA)}, pages 3022--3028. IEEE, 2023{\natexlab{a}}.

\bibitem[Zhang et~al.(2023{\natexlab{b}})Zhang, Ghimire, Zhang, Xu, and
  Ren]{zhang2023value}
Lei Zhang, Mukesh Ghimire, Wenlong Zhang, Zhe Xu, and Yi~Ren.
\newblock Value approximation for two-player general-sum differential games
  with state constraints, 2023{\natexlab{b}}.

\bibitem[Zhang et~al.(2018)Zhang, Wang, and Chen]{zhang2018continuous}
Wenzhao Zhang, Binfu Wang, and Dewang Chen.
\newblock Continuous-time constrained stochastic games with average criteria.
\newblock \emph{Operations Research Letters}, 46\penalty0 (1):\penalty0
  109--115, 2018.

\end{thebibliography}


\end{document}